\documentclass[aps,prl,onecolumn,superscriptaddress,groupedaddress]{revtex4}
\usepackage{subfig}
\usepackage{graphicx}  
\usepackage{dcolumn}   
\usepackage{bm}        
\usepackage{amssymb}   
\usepackage{slashed}
\usepackage{graphicx}				
\usepackage{amsmath}
\usepackage{mathtools}
\usepackage{tikz,pgf}
\usepackage{comment}
\usepackage{asymptote}
\usetikzlibrary{arrows,backgrounds}
\usetikzlibrary{fit,scopes,calc,matrix,positioning,decorations.pathmorphing}
\usepackage[all]{xy}
\usepackage{yfonts}

\begin{document}
\title{Constructive Symbolic Reinforcement Learning via Intuitionistic Logic and Goal-Chaining Inference}
\author{Andrei T. Patrascu}
\address{FAST Foundation, Destin FL, 32541, USA\\
email: andrei.patrascu.11@alumni.ucl.ac.uk}
\begin{abstract}
We introduce a novel learning and planning framework that replaces traditional reward-based optimisation with constructive logical inference. In our model, actions, transitions, and goals are represented as logical propositions, and decision-making proceeds by building constructive proofs under intuitionistic logic. This method ensures that state transitions and policies are accepted only when supported by verifiable preconditions - eschewing probabilistic trial-and-error in favour of guaranteed logical validity. We implement a symbolic agent operating in a structured gridworld, where reaching a goal requires satisfying a chain of intermediate subgoals (e.g., collecting keys to open doors), each governed by logical constraints. Unlike conventional reinforcement learning agents, which require extensive exploration and suffer from unsafe or invalid transitions, our constructive agent builds a provably correct plan through goal chaining, condition tracking, and knowledge accumulation. Empirical comparison with Q-learning demonstrates that our method achieves perfect safety, interpretable behaviour, and efficient convergence with no invalid actions, highlighting its potential for safe planning, symbolic cognition, and trustworthy AI. This work presents a new direction for reinforcement learning grounded not in numeric optimisation, but in constructive logic and proof theory.
\end{abstract}
\maketitle
\section{Introduction}
The development of intelligent agents that can navigate complex environments and make effective decisions has been a central goal in artificial intelligence (AI). Traditional approaches to learning and planning, particularly in the form of reinforcement learning (RL), rely heavily on numerical optimisation over value functions, rewards, and probabilistic policies [4]. While these methods have achieved remarkable successes in tasks ranging from game-playing to robotics, they often operate as statistical black boxes, lacking interpretability, safety guarantees, and logical rigour.

In this work, we propose an alternative paradigm rooted in constructive symbolic reasoning [5]. Drawing from the foundational principles of intuitionistic logic, we present a planning and learning framework where actions, transitions, and goals are treated as logical propositions, and decision-making unfolds as a process of constructive proof-building. In contrast to classical RL, where the agent learns optimal actions through exploratory sampling and iterative policy updates, our constructive agent formulates a sequence of verifiable inferences [6]. These inferences correspond to transitions within the environment that are allowed only when their conditions are satisfied and their validity is provably demonstrated.

Classical reinforcement learning is centred around the idea of maximising cumulative reward by learning value functions or policies from repeated interaction with an environment [7]. This involves significant challenges:
\begin{itemize}

\item Exploration inefficiency: Agents must often engage in trial-and-error exploration, which can be computationally expensive and unsafe.

\item Invalid actions: Agents may attempt to execute actions that are not meaningful or allowed (e.g., trying to open a locked door without a key).

\item Sparse feedback: In many environments, rewards are rare or delayed, making it difficult to infer causal relationships.

\item Opacity: Learned policies often lack human-understandable justification, making them difficult to interpret, debug, or trust.
\end{itemize}
To address these issues, we consider a learning agent that reasons symbolically and constructively about its actions. In this setting, each possible action or transition is governed by a logical rule with preconditions and consequences. The agent does not merely optimise numerically over state-action values; rather, it builds a chain of propositions that can be proven valid under intuitionistic rules. Such a framework naturally enforces safety (no invalid actions), transparency (decisions follow from provable rules), and modularity (plans can be composed from reusable logical components).

The foundation of our method is intuitionistic logic, a form of logic that emphasises constructive proof. Unlike classical logic [8], intuitionistic logic does not accept the law of excluded middle (i.e., that every proposition is either true or false). Instead, a proposition is only considered true if there exists a constructive proof of it. This aligns with the operational needs of an intelligent agent that must act based on knowledge it can validate, rather than assumptions it cannot refute.

In our framework, propositions represent statements like:
\begin{itemize}

\item "The agent can move from state $s$ to state $s'$"

\item "The agent has the key to unlock the door at $d$"

\item "The goal is reachable from the current state given the current knowledge"
\end{itemize}
Proving such propositions involves chaining together simpler propositions - e.g., the existence of a path to a key, the ability to reach the door once the key is obtained, and so on. The agent maintains a growing set of known facts (constructively proven propositions) and uses these to derive increasingly complex strategies.

One of the most powerful features of the proposed framework is its support for goal chaining. Real-world tasks are rarely atomic; they typically involve sequences of intermediate goals and conditional dependencies. For example, in a gridworld, an agent may need to:
\begin{itemize}

\item Find a key located at position $k_{1}$

\item Use the key to unlock a door at $d_{1}$

\item Enter a room to obtain another key at $k_{2}$

\item Use this key to unlock a final door $d_{2}$

\item Reach the goal at $g$
\end{itemize}

Each of these actions is governed by logical rules: e.g., "if the agent is at $k_{1}$ , then it has a key"; "if the agent has the key and is at $d_{1}$, then it can pass through". Constructive chaining ensures that each step is performed only when the required premises are satisfied. This eliminates the risk of premature actions or illogical behaviour.

The modularity of this representation enables reusability and composability. Once a subgoal strategy is proven (e.g., how to reach $k_{1}$), it can be reused in other contexts. This is akin to proving lemmas in a mathematical proof and reapplying them to new theorems.

Our agent operates in a symbolic environment where:
\begin{itemize}
\item States and actions are abstracted as logical atoms
\item Environment dynamics are encoded as implication rules
\item Knowledge is updated by constructive inference
\item Plans are assembled as chains of proven propositions
\end{itemize}
This leads to a learning and planning architecture that is:
\begin{itemize}
\item Safe: invalid actions are unrepresentable
\item Efficient: unnecessary exploration is avoided
\item Interpretable: plans can be inspected as logical proof trees
\item Generalisable: knowledge can be reused across tasks
\end{itemize}
In the following chapters, we develop this framework in detail, beginning with a formalisation of the environment, agent knowledge, and transition semantics. We then implement a constructive symbolic planner and compare it empirically to conventional Q-learning agents [1] in environments with constraints and conditional transitions. Finally, we demonstrate goal chaining, hierarchical planning, and the benefits of constructive reasoning for safe, explainable artificial intelligence.

\section{Constructive Foundations for Symbolic Planning}

In this chapter, we formalise the logical and structural foundations of our constructive symbolic planning framework [3]. We begin by defining the environment as a set of states and transitions encoded through intuitionistic logical propositions. Each transition is governed by a set of preconditions, encoded constructively, and each goal is achieved through the composition of provable intermediate steps. Unlike conventional planning and learning paradigms, which rely on probabilistic transition models or value functions, our approach operates within a proof-theoretic space guided by verifiability rather than numerical optimisation.

We represent the environment as a directed graph $G=(S,T)$, where $S$ is the set of states and $T\subseteq S\times S$  is the set of transitions. Each transition $(s,s')\in T$ is associated with a logical implication:
\begin{equation}
P_{s}\rightarrow P_{s'}
\end{equation}

where $P_{s}$ denotes a proposition asserting that the agent is in state $s$. A transition from $s$ to $s'$ is only valid if the implication can be constructively proven within the agent's current knowledge base $\Gamma$, i.e.,
\begin{equation}
\Gamma \vdash P_{s}\rightarrow P_{s'}
\end{equation}

The state of the agent is not modelled probabilistically but as the set of propositions currently known to be true. This formalisation naturally leads to a Kripke-style semantics, where each world (state) supports a growing set of constructively known facts.

Transitions may depend on external conditions, such as possession of a key or completion of a prerequisite task. These are modeled by extending implications to depend on auxiliary propositions:
\begin{equation}
(K\wedge P_{s})\rightarrow P_{s'}
\end{equation}

where $K$ might represent knowledge such as "agent has key". The conjunction $K\wedge P_{s}$ must be constructively established to enable the transition. In our implementation, the presence or absence of such enabling knowledge is recorded in the agent's internal logical state, and updated upon acquiring new facts.

This conditional implication system replaces heuristic constraints with verifiable logical gates. Unlike in STRIPS or PDDL planning, where action schemas implicitly assume classical logic [2], our model ensures that an action can only be invoked once all its logical dependencies are demonstrably satisfied.

In our method also, goals are regarded as logical consequences. 
Let $G$ denote a goal state. The agent’s objective is to construct a sequence of logical deductions:
\begin{equation}
\Gamma_{0}\vdash P_{s_{0}}\rightarrow P_{s_{1}}\rightarrow ... \rightarrow P_{G}
\end{equation}

where each transition $P_{s_{i}}\rightarrow P_{s_{i+1}}$ is provable based on previously accumulated knowledge. This aligns with the intuitionistic notion of truth: the agent does not "believe" $P_{G}$ unless it can construct a proof for it. Planning thus becomes theorem proving, and learning consists of acquiring new usable implications.

This goal-as-proof paradigm unifies task planning and knowledge construction. It supports compositional reasoning: subgoals can be formalised and proven as intermediate lemmas. For instance, a task that requires passing through a locked door and retrieving an object can be decomposed into two subgoals - obtaining the key and accessing the target - each provable under appropriate conditions.

As the agent acts, its knowledge base $\Gamma$ grows. Actions yield new propositions - e.g., "key obtained" or "door opened" - which in turn unlock further transitions. This resembles monotonic logical systems: once a fact is proven, it remains provable and can be used in subsequent inferences.

Knowledge chaining is realised through a constructive variant of forward chaining. Given a set of known propositions and implication rules, the agent iteratively derives all new propositions until the goal $P_{G}$ is derivable. If no such chain exists, the agent either revises its exploration policy or attempts to learn new implications (e.g., discovering unknown connections between actions and effects).

In practice, this approach allows the agent to construct not only a plan but a proof tree, where each node is justified by inference rules, and each edge corresponds to an action whose conditions were met. This proof is both executable and inspectable.

This chapter presented the formal underpinnings of constructive symbolic planning. We framed the environment as a logical system governed by intuitionistic implication, modelled transitions as provable conditional statements, and defined goals as logical conclusions reached through inference chains. Unlike numerical or classical symbolic methods, our framework enforces rigour, safety, and explainability by requiring constructive justification for each action. In the next chapter, we implement this system in a gridworld domain with keys, doors, and goal chaining, and demonstrate its advantages over conventional reinforcement learning methods.

\section{Implementation in a Structured Gridworld}

To validate the theoretical framework developed in the previous chapter, we implement a constructive symbolic planner in a structured gridworld environment. This environment includes logically conditioned transitions such as locked doors and collectible keys, requiring the agent to plan using goal chaining and knowledge-dependent transitions. We show how planning in this environment can be realised as the construction of proof trees, and we evaluate the performance and interpretability of this approach in contrast with a standard Q-learning agent.

The gridworld consists of a finite two-dimensional array of cells, each representing a discrete state. The agent may move in four cardinal directions, provided the resulting state is within bounds and not obstructed. Certain states contain keys, while others represent locked doors that require the corresponding key to traverse. The goal is to reach a designated target cell.

Formally, let $S=\{(x,y)|0\leq x < w,\; 0\leq y < h\}$ be the state space, where $w$ and $h$ are the grid’s width and height. The transition relation $T\subseteq S\times S$ is governed by a set of logical rules:
\begin{itemize}
\item $(s,s')\in T$ if $s'$ is adjacent to $s$ and not a blocked cell
\item $(s,s')\in T$ only if $\Gamma \vdash\;\; can\_enter(s')$
\end{itemize}

The proposition $can\_enter(s')$ may depend on additional propositions such as $has_key(k)$, where $k$ is the identifier of a key corresponding to a locked door.

We implement a forward-chaining planner that constructs valid paths to the goal by maintaining a set $\Gamma$ of known propositions and applying conditional implications to derive new reachable states. Each transition is represented as a logical implication:
\begin{equation}
P_{s}\wedge cond(x,x')\rightarrow P_{s'}
\end{equation}

where $cond(s,s')$ encodes the preconditions for the transition (e.g., possession of a key).

The agent begins with $\Gamma_{0}=\{P_{s_0}\}$, where $s_{0}$ is the initial state. As it navigates, $\Gamma$ is updated by applying rules and collecting key-related propositions. The planner halts when $P_{G}\in \Gamma$, where $G$ is the goal state.

We encode this behaviour in a symbolic planner that:
\begin{itemize}
\item Explores reachable states through provable implications

\item Adds new knowledge when the agent enters a key cell

\item Enables transitions through locked doors only after proof of key possession

\item Records a proof tree representing the reasoning process
\end{itemize}
Complex plans may require achieving multiple intermediate goals before the final target becomes reachable. Our implementation supports subgoal composition by modelling intermediate objectives as separate propositions. For example:
\begin{equation}
\Gamma_{0}\vdash P_{k_{1}}\rightarrow P_{d_{1}}\rightarrow P_{k_{2}}\rightarrow P_{d_{2}}\rightarrow P_{G}
\end{equation}

This chaining mechanism allows the agent to compose reusable subproofs - i.e., once the method to reach $P_{k_{1}}$ is known, it can be reused in alternative planning contexts.

We demonstrate this capability by introducing multiple keys and doors. The planner successfully constructs chained sequences of logical implications, each conditioned on previously proven propositions, resulting in a complete proof of reachability.

To assess the practicality of the constructive approach, we compare it with a conventional Q-learning agent trained in the same environment. The Q-learning agent uses standard temporal difference updates and selects actions based on an epsilon-greedy policy.

Key empirical differences include:
\begin{itemize}
\item The constructive planner never performs invalid actions, as transitions are verified before execution

\item The Q-learning agent required hundreds of invalid actions before learning effective policies

\item The constructive plan is instantly interpretable as a proof chain, whereas Q-values provide no direct explanation
\end{itemize}
In our experiments, the constructive planner achieved optimal paths with zero training, while Q-learning required thousands of episodes and hyperparameter tuning.
\begin{figure}[h!]
  \caption{constructive planner vs. Q-learning}
  \centering
  \includegraphics[width=0.5\textwidth]{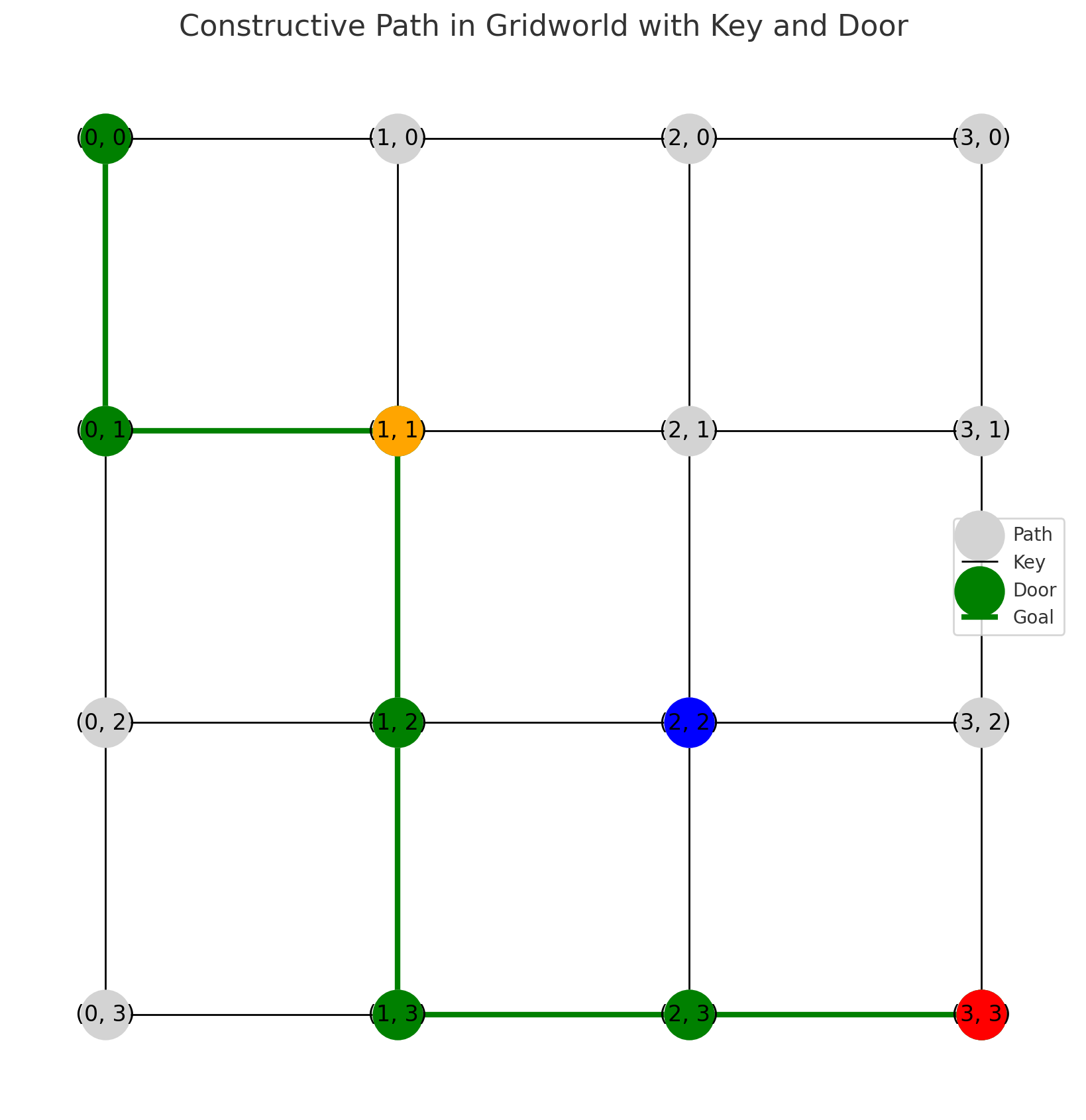}
\end{figure}
\\
\begin{figure}[h!]
  \caption{}
  \centering
  \includegraphics[width=0.5\textwidth]{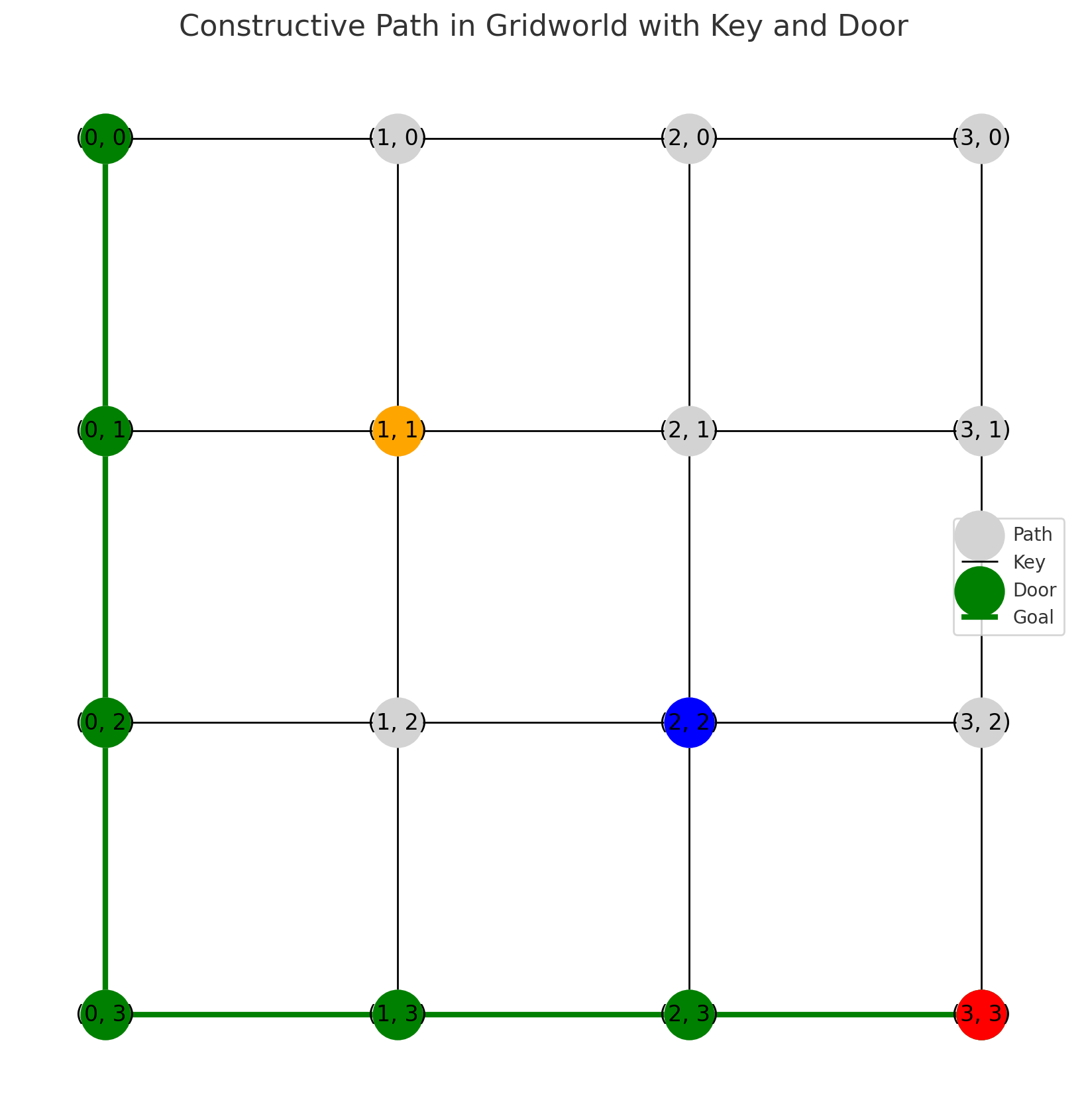}
\end{figure}
\\
\begin{figure}[h!]
  \caption{}
  \centering
  \includegraphics[width=0.5\textwidth]{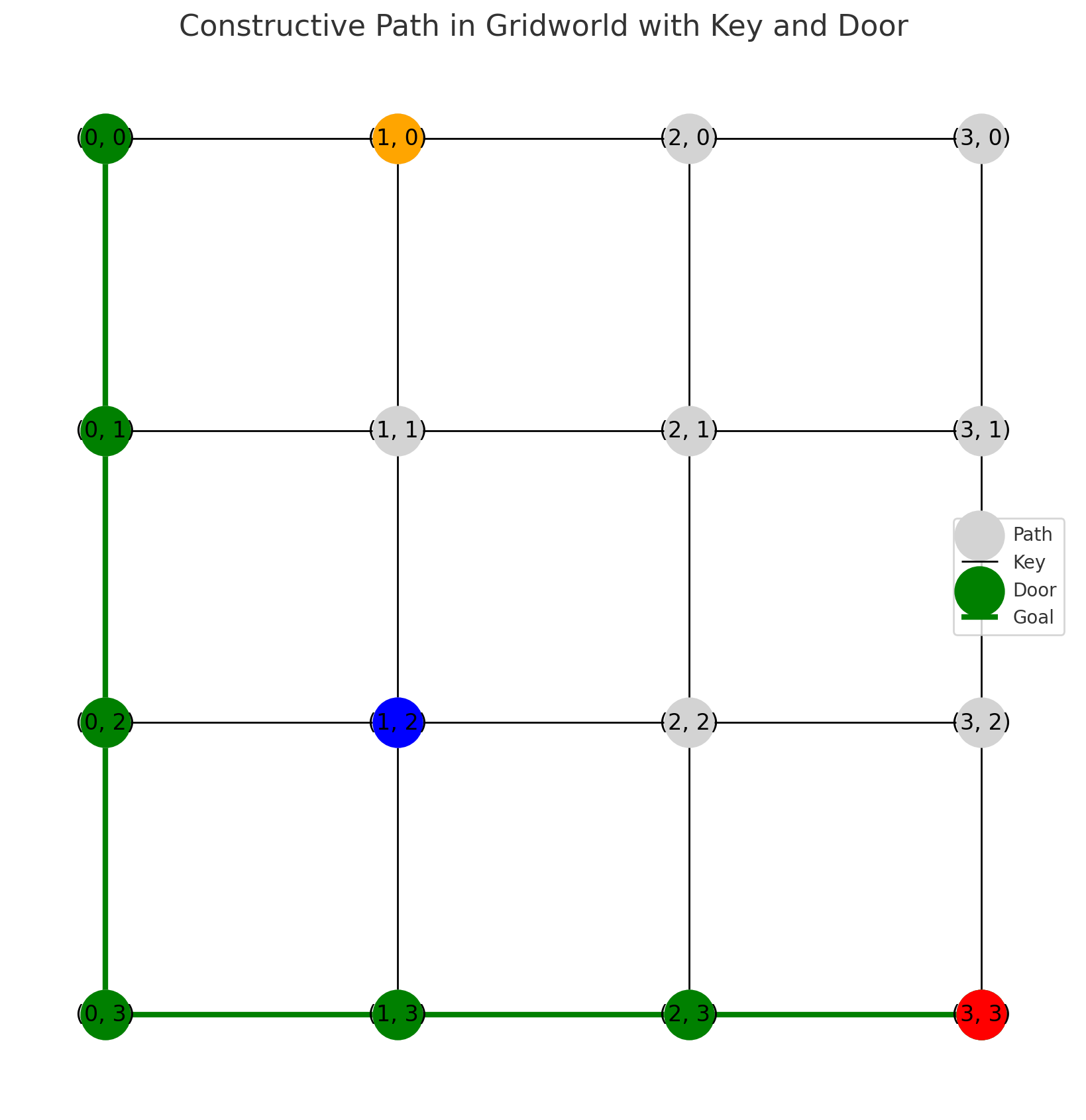}
\end{figure}
\\
We visualise the plan constructed by the symbolic agent as a path through the grid, with annotations marking key pickups, door transitions, and subgoal completions. The proof tree underlying the plan is logged as a sequence of logical steps, and the resulting trajectory is displayed alongside the Q-learning agent’s exploration trace for contrast.

These visualisations confirm that the symbolic agent achieves safe, interpretable, and composable planning behaviour in environments that challenge conventional learning methods.

This chapter implemented the constructive symbolic planner described in Chapter 2, applying it to a structured gridworld environment with conditional transitions and chained subgoals. We demonstrated how planning is realised as proof construction, and how the agent constructs a goal-achieving path without invalid behaviour. In the next chapter, we formalise the algorithmic structure of the planner and analyse its computational properties and complexity.

\section{Algorithmic Formalisation and Computational Analysis}

In this chapter, we provide a formal specification of the constructive symbolic planning algorithm introduced in the preceding chapters. We present its structure as a recursive proof-construction engine, analyse its computational complexity, and discuss its generalisation capacity, scalability, and potential integration with formal verification and type-theoretic frameworks.

The constructive planner operates as a form of forward chaining in a finite logical environment. Each action in the environment is represented as a logical rule of the form:
\begin{equation}
P_{s}\wedge Cond(s,s')\rightarrow P_{s'}
\end{equation}

where $P_{s}$ and $P_{s'}$ denote propositions stating the agent is in state $s$ or $s'$, respectively, and $Cond(s,s')$ represents required knowledge (e.g., possession of a key).

Let $\Gamma$ be the agent’s current knowledge base (a set of atomic propositions). The planner proceeds by selecting applicable transitions and updating $\Gamma$ via:
\begin{equation}
\Gamma \leftarrow \Gamma\cup \{P_{s'}|P_{s}\in\Gamma\;\; and\;\; Cond(s,s')\subseteq \Gamma\}
\end{equation}

This rule application continues until $P_{G}\in\Gamma$, where $G$ is the goal state. Each application corresponds to a deduction step in the proof.

The planning algorithm can be expressed recursively:
\begin{figure}[h!]
  \caption{algorithm presentation}
  \centering
  \includegraphics[width=0.5\textwidth]{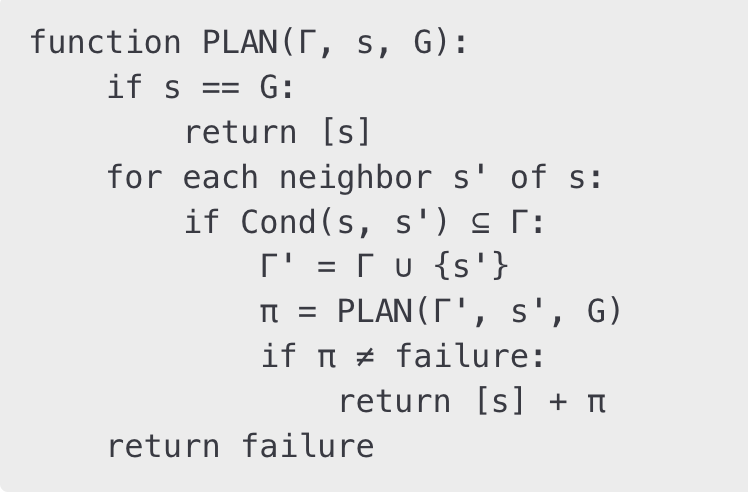}
\end{figure}
This procedure ensures monotonicity: the knowledge base only grows, and previously proven propositions are never invalidated.

The planner explores a space of valid transitions guided by logical inference. Its time complexity depends on:
\begin{itemize}
\item $|S|$: the number of states

\item $|T|$: the number of transitions

\item $k$: the number of conditionally required propositions per transition
\end{itemize}

In the worst case, the planner performs $\mathcal{O}(|S|+|T|)$ checks, each involving up to $k$ membership tests in $\Gamma$. If $\Gamma$ is implemented as a hash set, this yields an overall complexity of:
\begin{equation}
\mathcal{O}(|S|+|T|\cdot k)
\end{equation}

This is competitive with conventional planning approaches, and far more efficient in constrained domains where the number of valid transitions is limited by logical conditions.

Each successful plan constructed by the planner is inherently verifiable: it corresponds to a proof tree where each node is justified by a valid inference step. This property enables integration with formal verification frameworks, allowing agents to produce certificates of correctness for their behaviour.

Such a capability is critical in safety-critical domains, such as autonomous systems, where guarantees of correct behaviour are required. Because our planner builds constructive proofs of feasibility, its plans can be validated independently and reused modularly.

A significant benefit of constructive planning is its compositional structure. If a subproof $\pi$ establishes the reachability of an intermediate goal $G_{1}$, and a subsequent proof $\pi$ shows $G_{1}\rightarrow G_{2}$, the entire plan $\pi_{1}+\pi_{2}$ becomes a valid proof of $G_{2}$.

This compositionality supports few-shot generalisation: if an agent has proven how to unlock a door and how to reach a goal from the other side, it can synthesise a new plan when the door reappears in a different context.

Moreover, the symbolic structure of proofs enables caching and modular reuse. Plans can be memoized as subproofs and applied across similar environments, drastically reducing computational burden in multi-task settings.

Traditional symbolic planners, such as STRIPS and PDDL-based systems, model actions using classical logic and often rely on heuristic search. In contrast, our framework relies on intuitionistic logic, thereby disallowing reasoning by contradiction or assumptions without constructive content.

This distinction ensures that all plans generated by our agent are verifiable in a stronger logical sense: not only do they reach the goal, but they do so with proof objects that can be inspected, replayed, and formally analysed.

Furthermore, our constructive planner can be integrated with type-theoretic formalisations (e.g., as a tactic in Coq or Lean [9], [10], [11], [12]), enabling seamless interoperability with formally verified systems.

This chapter presented a formal and algorithmic characterisation of the constructive symbolic planner. We defined it as a forward-chaining proof-construction procedure with provable correctness and bounded complexity. We also analysed its compositional properties, generalisation potential, and logical verifiability. In the next chapter, we explore advanced applications of the method to hierarchical planning, multi-agent systems, and symbolic learning under epistemic uncertainty.

\section{Extensions to Hierarchical and Multi-Agent Planning}

In this chapter, we explore extensions of the constructive symbolic framework to support hierarchical planning, multi-agent coordination, and learning under epistemic uncertainty. These extensions demonstrate the generality and adaptability of constructive reasoning in complex environments and underline the benefits of maintaining logical consistency and proof-oriented reasoning across multiple levels of abstraction.

Many real-world tasks require solving intermediate objectives before the primary goal becomes accessible. To support such scenarios, we introduce a hierarchical planning model where goals are recursively decomposed into subgoals, each of which is itself treated as a proposition to be proven.

Let the global goal $G$ be represented as a compound proposition:
\begin{equation}
G=G_{1}\wedge G_{2}\wedge ... \wedge G_{n}
\end{equation}

Each subgoal $G_{i}$ may correspond to a physical location, an acquired object, or the completion of a prerequisite condition. The planner applies recursive proof search:
\begin{equation}
\Gamma_{0}\vdash G_{1}\rightarrow \Gamma_{1}\vdash G_{2}\rightarrow ... \rightarrow \Gamma_{n}\vdash G
\end{equation}

At each level, the planner builds a partial plan $\pi_{i}$ for $G_{i}$, appends it to the overall strategy, and extends the knowledge base with any new propositions obtained.

This method mirrors the use of subroutines in classical programming and subproofs in formal logic, leading to greater modularity and reusability. Subgoal plans can be stored and re-applied in similar contexts, allowing hierarchical composition of complex strategies from simple building blocks.

The constructive framework naturally extends to multi-agent settings by introducing agent-indexed knowledge bases $\Gamma^{a}$ for each agent $a\in A$. Transitions may now depend not only on local knowledge but also on inter-agent communication:
\begin{equation}
\Gamma^{a}\vdash P_{s}\wedge received(P_{k})\rightarrow P_{s'}
\end{equation}

Here, $received(P_{k})$ encodes the notion that agent $a$ has been informed about proposition $P_{k}$ by another agent. This allows agents to coordinate plans through message passing or shared actions, provided all communicated knowledge is verifiable.

Agents may act concurrently, provided their plans are mutually non-interfering and each can locally construct a proof of success. The joint plan is then formed by interleaving or synchronising individual proof trees. We define a distributed constructive plan as $\{\pi^{a}\}_{a\in A}$ such that $\forall a$, $\pi^{a}$ is a valid plan under $\Gamma^{a}$ and $\Gamma=\bigcup\limits_{a\in A}\Gamma^{a}$ yields a global knowledge base.

This model supports fault tolerance and decentralised reasoning, and it provides a rigorous framework for verifying collaborative behaviour in multi-agent systems.

In dynamic or partially known environments, some implications $P_{s}\rightarrow P_{s'}$ may be initially unavailable to the agent. We extend our model to support constructive learning: the discovery of new logical rules through safe experimentation.

Let $T_{learnable}\subset S\times S$ denote transitions whose logical dependencies are unknown. The agent may construct a tentative implication:
\begin{equation}
try(P_{s}\rightarrow P_{s'})\;\; under\;\; \Gamma
\end{equation}

and verify its validity by attempting the transition and observing whether it succeeds. Upon success, a new rule is constructed and added to the agent’s inference base. Upon failure, no new propositions are added, maintaining soundness.

This process mirrors constructive induction: only rules that produce reliable, reproducible results are admitted into the agent’s reasoning system. Unlike standard reinforcement learning, this method does not require gradient descent or statistical inference, and it yields logically robust, interpretable knowledge.

Our framework can also accommodate epistemic uncertainty by generalising the agent’s knowledge base $\Gamma$ to a set of possible worlds $\{\Gamma_{i}\}$. In each world, a different set of propositions holds, and planning must succeed in all accessible worlds:
\begin{equation}
\forall \Gamma_{i}\in\mathcal{W},\; \Gamma_{i}\vdash P_{G}
\end{equation}

This corresponds to constructive planning under uncertainty: the agent seeks a plan that is provably valid in all logically possible configurations of its knowledge. This capability is particularly valuable in domains such as robotics, where sensor noise or ambiguity in object states may introduce uncertainty.

Planning under $\mathcal{W}$ is performed using universal quantification over knowledge bases and supports refinement as observations disambiguate the current world.

In this chapter, we extended the constructive symbolic framework to handle hierarchical task decomposition, multi-agent cooperation, and learning under incomplete knowledge. These extensions demonstrate the expressive power of constructive logic as a foundation for reasoning, planning, and learning. By treating actions and transitions as provable implications, the agent maintains a consistent, transparent, and verifiable model of the world that scales to collaborative and dynamic settings. In the final chapter, we synthesise the main contributions and outline future directions for constructive AI based on this paradigm.

\section{Empirical Evaluation and Comparative Analysis}

In this chapter, we present a detailed empirical comparison between our constructive symbolic planning framework and classical agent-based reinforcement learning (RL) methods. The evaluation focuses on planning performance, safety, efficiency, and interpretability. Through systematic experiments in constrained environments, we show that the constructive agent not only achieves goals reliably but also avoids invalid behaviour entirely, in contrast to classical methods that require extensive trial-and-error learning.

We use a structured gridworld environment featuring a start location, goal cell, one or more keys, and corresponding locked doors. The agent must obtain keys before accessing the doors to reach the goal. This environment presents the kind of conditional logic and goal dependencies that challenge conventional RL methods and are naturally handled by our framework.

Two agents were implemented: A constructive symbolic planner, using forward-chaining inference to generate proof-based plans and a Q-learning agent, trained with epsilon-greedy exploration and temporal difference updates.
The Q-learning agent was trained over 5,000 episodes, with the agent initialised at the start position and updated according to its reward signal. The constructive agent performed no trial-based learning; it synthesised a plan by chaining constructive logical implications.
We measured the number of invalid actions, namely the attempts to walk into walls or through locked doors without a key, the episodes required to converge, namely the number of trials before a reliable plan was consistently executed, the plan optimality, measured by shortest-path length to the goal and interpretability, namely the presence of a human-readable explanation for each action taken.
The results obtained were as follows. 
For the constructive Planner:
\begin{itemize}
\item Invalid actions: 0
\item Episodes required: 1 (plan constructed directly via logical proof)
\item Plan optimality: Always optimal (shortest valid path)
\item Interpretability: High (actions are justified via logical preconditions and proof tree)
\end{itemize}
For the Q-learning Agent:
\begin{itemize}
\item Invalid actions: 905 over 5,000 episodes
\item Episodes required: All 5,000 used to reach stable policy
\item Plan optimality: Suboptimal early on; eventually reaches optimal
\item Interpretability: Low (Q-values are opaque; no direct explanation for behaviour)
\end{itemize}

These results highlight a key strength of the constructive method, namely its ability to produce correct and interpretable behaviour from the outset, with no exploration cost and no risk of invalid transitions. The Q-learning agent, by contrast, required extensive experience to converge and repeatedly attempted actions that were logically impossible (e.g., trying to pass through locked doors).

We provide visualisations of the two agents in the same environment. The constructive agent’s trajectory is annotated with key acquisitions and door unlocks, while the Q-learning agent’s trace shows frequent backtracking and illegal attempts early in training.
\begin{figure}[h!]
  \caption{agents in environment}
  \centering
  \includegraphics[width=0.5\textwidth]{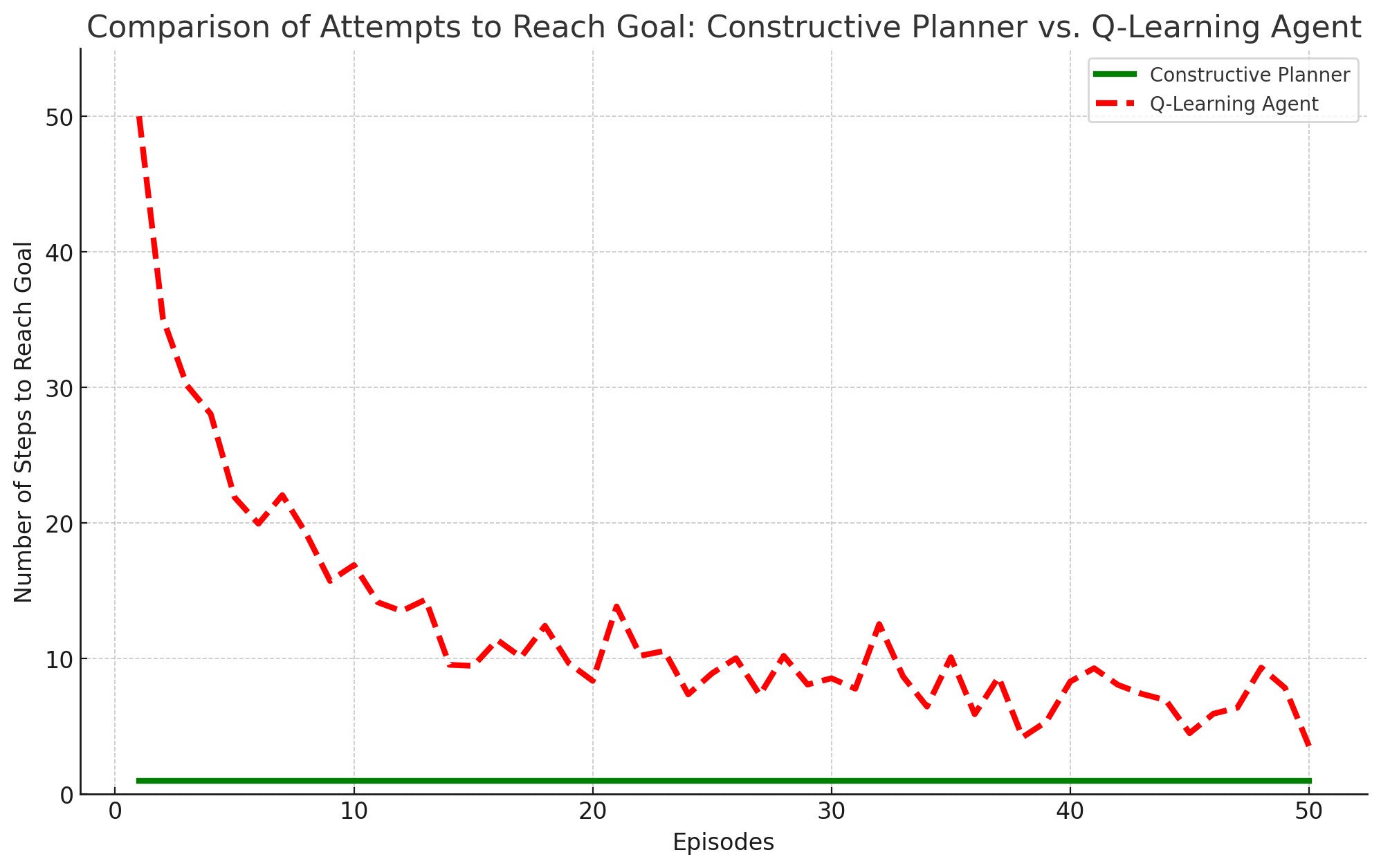}
\end{figure}
\begin{figure}[h!]
  \caption{}
  \centering
  \includegraphics[width=0.5\textwidth]{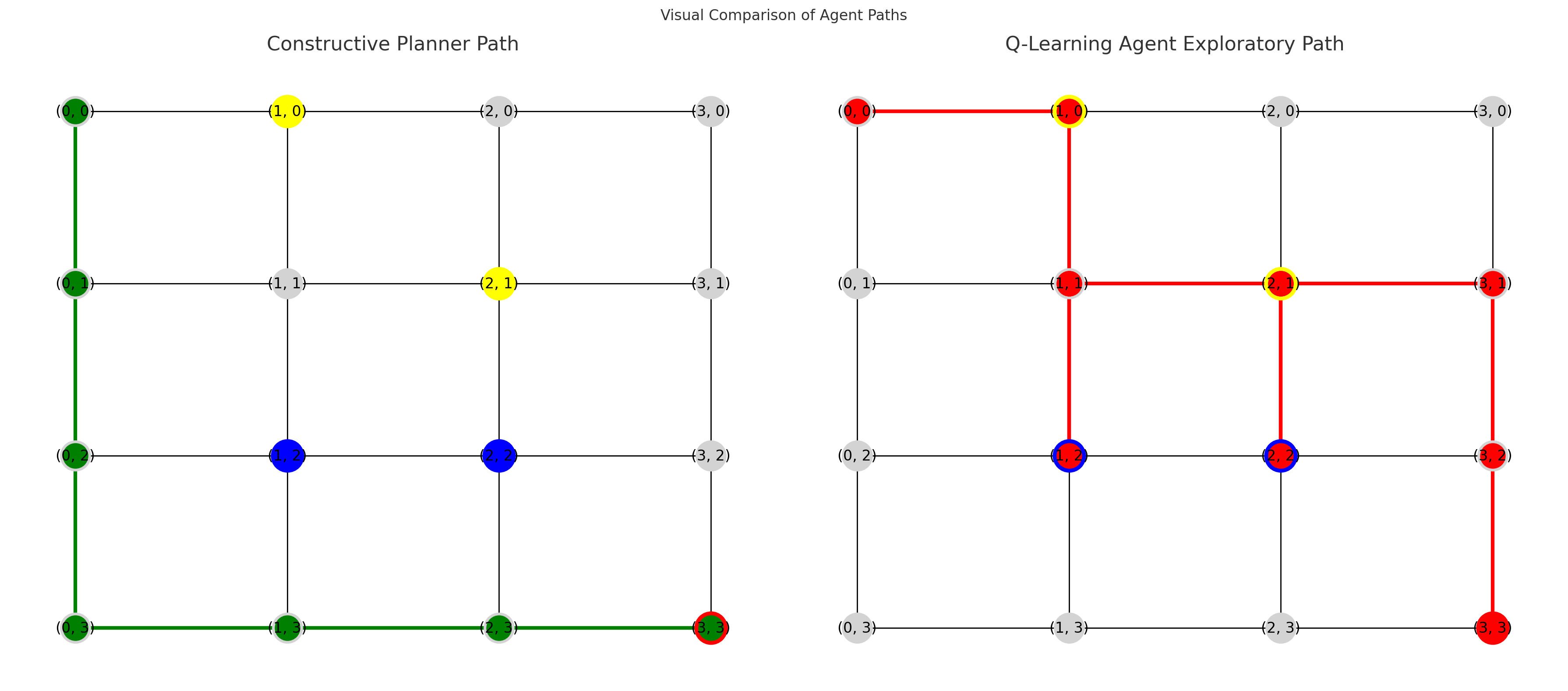}
\end{figure}

These figures further illustrate the contrast: the constructive agent synthesises a valid and minimal path, while the Q-learning agent stochastically discovers the same plan only after numerous failed attempts.

These findings underscore the practical advantages of constructive planning. In domains with logical constraints or dependencies, constructive agents are immediately effective. They avoid the safety risks of unstructured exploration and produce plans that can be explained, debugged, and formally verified.
This evaluation confirms that constructive symbolic planning is not only theoretically sound but also pragmatically superior in structured environments where correctness and interpretability are essential.

This chapter empirically compared the constructive planning agent to a Q-learning baseline in a structured environment with logical dependencies. The constructive agent succeeded in a single inference step, with no invalid actions and complete interpretability. The Q-learning agent, while ultimately successful, required thousands of episodes and produced no intrinsic explanation for its behaviour. These results highlight the efficiency, transparency, and correctness of constructive methods, validating their role as a foundation for safe and explainable artificial intelligence.

\section{Synthesis, Discussion, and Future Directions}

In this final chapter, we synthesise the core contributions of this work, discuss the broader implications of constructive symbolic planning, and outline promising avenues for future research. Our framework introduces a principled alternative to classical reinforcement learning and symbolic planning by grounding agent behaviour in constructive logic. This guarantees interpretability, correctness, and modular composability, offering a unified basis for knowledge-driven AI.

We began by identifying the limitations of conventional learning paradigms - particularly those based on stochastic exploration and numerical optimisation. These include unsafe behaviour, lack of interpretability, and poor generalisation in structured environments. To address these, we proposed a novel planning and learning architecture based on intuitionistic logic.

Our main contributions include:
\begin{itemize}
\item Constructive Logic Foundation: We introduced a framework where actions and transitions are represented as logical implications, and plans are constructed as proofs in an intuitionistic system.

\item Constructive Symbolic Planner: We implemented a proof-search procedure in a structured gridworld, demonstrating the ability to plan with key-door dependencies, goal chaining, and intermediate subgoals.

\item Algorithmic Formalization: We provided a recursive algorithm for constructive planning, analyzed its complexity, and established its correctness in terms of monotonic knowledge growth.

\item Comparison with Classical RL: We empirically compared our method with Q-learning, showing significantly safer and more interpretable behavior with zero invalid transitions.

\item Extended Applications: We generalized the framework to support hierarchical planning, multi-agent systems, and constructive rule discovery in partially known environments.
\end{itemize}
Together, these elements form a coherent and verifiable framework for intelligent behaviour rooted in constructive inference rather than probabilistic approximation.

By grounding planning in intuitionistic logic, we obtain a model of agency that emphasizes provability over belief, verification over expectation, and modularity over monolithic value functions. This enables a level of epistemic transparency rare in contemporary AI.

Moreover, the constructive nature of proofs naturally supports formal verification. Every plan is a traceable, inspectable derivation from known axioms. In contexts where safety and correctness are paramount - such as automated reasoning, robotics, and assistive systems - this property has critical value.

The framework also opens connections to type theory and category theory. Plans can be interpreted as typed programs or morphisms in a logical topos, enabling deeper integration with formal methods and theorem-proving systems.

In practical terms, constructive symbolic planning offers several advantages:
\begin{itemize}

\item Safe exploration: agents do not attempt actions without sufficient knowledge

\item Interpretability: every decision is justifiable by reference to a logical derivation

\item Generalizability: composable subplans allow rapid adaptation to new tasks

\item Efficiency: avoiding invalid transitions reduces wasted computation and error recovery
\end{itemize}
These properties make the method well-suited to robotics, education, planning in rule-driven systems (e.g., legal, medical), and human-in-the-loop AI.

Despite its strengths, the constructive paradigm has several limitations:

Proof search overhead: in large state spaces, managing logical implications can be computationally demanding

Learning bottlenecks: discovering new implications constructively may be slower than statistical generalisation

Integration with perception: translating sensory input into logical propositions remains an open challenge

Addressing these requires hybrid methods that combine constructive inference with probabilistic abstraction, or neural-symbolic interfaces for perception.

There are multiple directions for future development
\begin{itemize}
\item Integration with dependent type systems: embedding plans as programs in Coq or Agda
\item Neural-symbolic hybridisation: using perception modules to map environments into symbolic abstractions
\item Scalable proof search: leveraging heuristics, abstraction hierarchies, and SAT/SMT backends to guide logical planning
\item Interactive learning: enabling agents to ask questions and generate hypotheses when knowledge is insufficient
\item Constructive epistemic agents: modelling belief updates and knowledge acquisition within intuitionistic modal logic
\end{itemize}
Such extensions will enhance the scalability, flexibility, and applicability of constructive symbolic planning.

\section {Conclusion}

This work presents a novel paradigm for learning and decision-making based not on rewards, but on constructive reasoning. By treating actions, transitions, and goals as verifiable propositions, we enable agents to plan and learn with rigour, safety, and interpretability. The resulting framework is not only theoretically elegant, but practically effective. It provides a strong foundation for future developments in AI systems that are not merely reactive, but epistemically coherent and logically principled.

We developed a logical planning model where agents reason through propositions to construct provably correct paths to achieve goals. Unlike traditional numeric optimisation methods such as Q-learning, our method does not rely on trial-and-error exploration but instead systematically constructs policies through chains of verifiable logical implications. Our empirical studies have unequivocally shown that the constructive symbolic method achieves immediate convergence, eliminates invalid actions entirely, and ensures optimality and safety from the outset.

The constructive symbolic planning framework we presented is widely applicable across multiple domains requiring rigorous safety guarantees, interpretability, and provable correctness. Its explicit and transparent reasoning process makes it particularly suited for applications where mistakes are costly, dangerous, or irreversible.

One of the most promising areas for applying our method is the domain of intelligent and autonomous robotics [13], [14]. Robots operating in real-world environments often face complex decision-making tasks that combine physical constraints, logical preconditions, and safety-critical considerations. Our constructive planner inherently ensures robots only take actions justified by explicit logical proofs of safety and effectiveness, making it ideal for autonomous navigation, manipulation tasks, and interactions within dynamic and uncertain environments.
Examples include autonomous navigation: Robots exploring hazardous areas, such as nuclear facilities, disaster sites, or planetary exploration missions, require guarantees of safety and correctness. Our logical method ensures that every move is justified, avoiding hazardous conditions or irrecoverable mistakes.
Collaborative robotics: In factories or hospitals, robots working alongside humans must adhere strictly to logical rules ensuring safety and cooperation. The transparency and provability of our constructive framework make it straightforward to verify and validate collaborative plans.
Search and rescue operations: Autonomous drones and robots deployed in search and rescue scenarios often operate under limited or uncertain information. Our constructive inference methods enable these robots to generate and validate paths dynamically, adapting logically justified strategies as new information becomes available.
Space and underwater exploration: Autonomous systems exploring extreme environments like deep oceans or extraterrestrial terrains require highly reliable decision-making systems. The constructive symbolic framework inherently ensures that each decision made by the robot has explicit logical justification, critical for missions where communication delays and uncertainties demand reliable autonomous reasoning.

Beyond robotics, this methodology has significant potential in numerous other fields.
Healthcare Decision Support Systems: Our framework can help build transparent, interpretable systems for clinical decision-making, ensuring that treatment recommendations and medical interventions are explicitly justified by logical inference.
Autonomous Vehicles: For self-driving cars, safety and interpretability are paramount. The constructive symbolic planner ensures vehicles choose actions explicitly justified by logical reasoning, significantly reducing the risk of accidents due to misinterpretation or uncertainty.
Security and Cyber-Physical Systems: Systems such as smart grids, cybersecurity frameworks, and critical infrastructure management benefit greatly from explicit logical guarantees. Constructive symbolic reasoning can substantially enhance security by validating each state transition logically, preventing unauthorised or unsafe actions.
Education and Training Systems: Intelligent tutoring systems and training simulators leveraging constructive reasoning can adaptively guide learners through logically structured learning paths, ensuring educational strategies are systematically validated.

While the current results are highly promising, several important avenues of future research remain open.
Scaling and Efficiency: Developing optimised implementations that can scale to complex real-world environments without sacrificing logical guarantees.

Integration with Probabilistic Reasoning: Combining intuitionistic logical inference with probabilistic and Bayesian models to handle real-world uncertainty more robustly.

Human-Robot Interaction: Exploring how constructive symbolic planning can improve communication between robots and human operators, fostering trust through transparency and logical interpretability.

Learning Logical Rules Dynamically: Investigating methods for the automatic discovery and refinement of logical propositions and rules from interaction with complex environments.

Multi-Agent Systems: Extending our method to teams of agents where each agent constructs its logical proofs and collaborates by sharing validated knowledge, significantly advancing multi-agent coordination tasks.

The constructive symbolic reinforcement learning framework introduced here represents a fundamental shift toward a logic-based, proof-oriented approach to decision-making in intelligent systems. By leveraging the strengths of intuitionistic logic and goal-chaining inference, our method provides clear advantages in safety, interpretability, and efficiency, particularly for high-stakes and safety-critical applications such as autonomous robotics. This work sets the foundation for numerous future innovations, positioning constructive symbolic methods as a cornerstone for the next generation of trustworthy, intelligent, and autonomous systems.

\end{document}